\documentclass[10pt, conference]{IEEEtran} 
\usepackage{cite}
\usepackage{subfigure}
\usepackage{amsmath,amssymb,amsfonts}
\usepackage{algorithmic}
\usepackage{graphicx}
\usepackage{multirow}
\usepackage{textcomp}
\def\BibTeX{{\rm B\kern-.05em{\sc i\kern-.025em b}\kern-.08em
    T\kern-.1667em\lower.7ex\hbox{E}\kern-.125emX}}
\begin{document}
%
\title{Offline Writer Identification based on the Path Signature Feature}

\author{
\IEEEauthorblockN{Songxuan Lai, Lianwen Jin}
\IEEEauthorblockA{\textit{School of Electronic and Information Engineering} \\
\textit{South China University of Technology}\\
Guangzhou, China\\
lai.sx@mail.scut.edu.cn, eelwjin@scut.edu.cn}
}


\maketitle

\begin{abstract}
In this paper, we propose a novel set of features for offline writer identification based on the path signature approach, which provides a principled way to express information contained in a path. By extracting local \emph{pathlets} from handwriting contours, the path signature can also characterize the offline handwriting style. A codebook method based on the log path signature---a more compact way to express the path signature---is used in this work and shows competitive results on several benchmark offline writer identification datasets, namely the IAM, Firemaker, CVL and ICDAR2013 writer identification contest dataset.
\end{abstract}


%
\IEEEpeerreviewmaketitle

\section{Introduction}
Offline writer identification is to determine the writer of a text among numerous known writers based on their handwriting images. With practical applications in forensic analysis, offline writer identification is an important research field in pattern recognition and has made progressive advances in the past decades \cite{plamondon1989automatic}\cite{bulacu2007text}\cite{he2017beyond}. Most existing approaches are text-independent and can be roughly divided into three categories: texture-based, structure-based and CNN-based approaches.

Texture-based approaches regard handwritten texts as a special texture image and extract texture features for writer identification. Bertolini et al. \cite{bertolini2013texture} used local binary patterns (LBP) and local phase quantization (LPQ) in the dissimilarity framework. He et al. \cite{he2017writer} designed the LBPruns method by carrying out binary tests on parallel scanning lines and counting the run lengths of the resulting LBP patterns. By considering the joint feature distribution (JFD) principle \cite{he2017beyond}, more powerful texture features, such as CoLBP, can be designed. Filter-based approaches, such as Gabor and XGabor filters \cite{helli2010text}, derivative-of-Gaussian filters \cite{newell2014writer}, and wavelet \cite{he2008writer}, are also commonly used in writer identification,

Structure-based approaches exploit the structural information of local handwriting traces, and can be divided into two main categories: contour-based and grapheme-based methods. The contour-based methods extract features from the contours of handwritings and capture the local geometric properties. Bulacu et al. \cite{bulacu2007text} proposed several contour-based directional features, of which the Hinge feature is considered to provide a strong baseline for following studies. Siddiqi et al. \cite{siddiqi2010text} used the global and local chain code-based features to capture orientation and curvature. Brink et al. \cite{brink2012writer} proposed the Quill feature which is a joint probability of the ink direction and the ink width. He et al. \cite{he2014delta} generalized the Hinge feature and introduced the rotation-invariant $\Delta^{n}$Hinge feature. A series of new contour-based features was introduced in \cite{he2017beyond} based on the JFD principle. In the grapheme-based methods, handwriting graphemes \cite{bulacu2007text} (obtained via segmentation) or patches \cite{siddiqi2010text} are used to generate a codebook that can capture the structural details of the allographs emitted by the writers. The SIFT \cite{wu2014offline}\cite{xiong2015text} and RootSIFT \cite{khan2019dissimilarity} are also very effective in offline writer identification, which can be viewed as scale-invariant graphemes in the Gaussian scale space.

CNN-based approaches train convolutional neural networks to extract discriminant features from handwriting patches \cite{christlein2014writer}, words \cite{he2019deep}, or pages \cite{tang2016text}. Although CNN achieves very promising results, it requires a heavy computation which limits its appeal in some scenarios.

In this paper, we propose a novel set of contour-based features based on the path signature (PS) approach. The PS was initially introduced in the rough paths theory as a branch of stochastic analysis and has recently been successfully applied to pattern recognition as a principled method for time series description, such as online handwriting recognition \cite{lai2017toward}\cite{xie2018learning}, online signature verification \cite{lai2017online}\cite{lai2018recurrent}, and skeleton-based action recognition \cite{yang2017leveraging}\cite{li2017lpsnet}. Although offline handwritten text is not a time series, by extracting the handwriting contours and segmenting them into local fragments, we can describe the analytic and geometric properties of such fragments with the PS. We denote such fragments as \emph{pathlets}. Compared with previous contour-based approaches, the PS method encodes rich information, such as orientation and curvature, in a principled way, and outperforms previous approaches on several benchmark datasets.

The rest of the paper is organized as follows. Section II introduces the PS theory in detail and shows how it can be used to characterize and identify offline handwritings. Section III reports the experimental results and analysis. Finally, Section IV concludes the paper.
\section{Methodology}
\begin{figure*}[t]
\centerline{\includegraphics[width=6.2in]{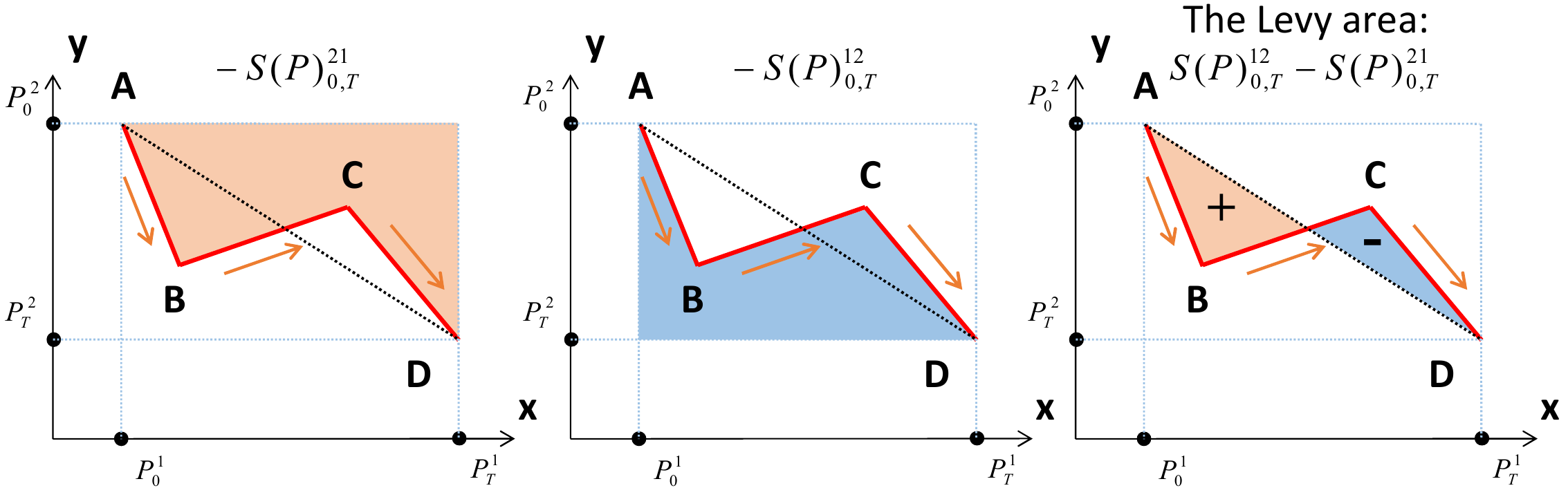}}
\caption{Geometric interpretations of $S(P)^{12}_{0, T}$ and $S(P)^{21}_{0, T}$. $S(P)^{12}_{0, T}$ and $S(P)^{21}_{0, T}$ are elements from the $2^{nd}$ level PS, while $S(P)^{12}_{0, T}-S(P)^{21}_{0, T}$ is an element from the $2^{nd}$ level LPS.}
\label{PS}
\end{figure*}
To make this paper more self-contained, in this section we first concisely introduce the PS and log path signature (LPS), and then show how we construct features based on the LPS for offline writer identification.
\subsection{Path Signature and Log Path Signature}
\subsubsection{Path Signature}
The information contained in a path can be expressed in the form of iterated integrals \cite{chevyrev2016primer}. Assume a path $P:[0, T]\mapsto \mathbb{R}^d$ where $[0, T]$ is a time interval. The coordinate paths are denoted by $\{P^{1},P^{2},...,P^{d}\}$, where each $P^i:[0, T]\mapsto \mathbb{R}$ is a real-valued path. The k-fold iterated integral of path $P$ \emph{along the indexes $i_1,...,i_k$} is defined as
\begin{equation}\label{Iter}
  S(P)^{i_{1},...,i_{k}}_{0, T}=\int_{0<t_1<...<t_k<T}dP^{i_{1}}_{t_1}...dP^{i_{k}}_{t_k},
\end{equation}
where $i_1,...,i_k\in\{1,...,d\}$, and $P^i_t$ denotes the value of path $P^i$ at time $t$ ($\in[0, T]$). For example, for a 2D path such as the handwriting trajectory, the 1-fold integrals have two terms:
\begin{equation}\label{inc}
\begin{split}
  &S(P)^{1}_{0, T}=\int_{0<t<T}dP^{1}_t=P^1_T-P^1_0,\\
  &S(P)^{2}_{0, T}=\int_{0<t<T}dP^{2}_t=P^2_T-P^2_0,
\end{split}
\end{equation}
which correspond to increments in the x, y directions, respectively. The 2-fold iterated integrals have four terms:
\begin{equation}\label{inc}
\begin{split}
  &S(P)^{11}_{0, T}=\int_{0<t_1<t_2<T}dP^{1}_{t_1}dP^{1}_{t_2}=\frac{(P^1_T-P^1_0)^2}{2},\\
  &S(P)^{22}_{0, T}=\int_{0<t_1<t_2<T}dP^{2}_{t_1}dP^{2}_{t_2}=\frac{(P^2_T-P^2_0)^2}{2},\\
  &S(P)^{12}_{0, T}=\int_{0<t_1<t_2<T}dP^{1}_{t_1}dP^{2}_{t_2},\\
  &S(P)^{21}_{0, T}=\int_{0<t_1<t_2<T}dP^{2}_{t_1}dP^{1}_{t_2},\\
\end{split}
\end{equation}
where the first two terms are proportional to the square of the corresponding increments, and the last two terms have intuitive geometric interpretation as shown in Fig. 1.

The signature of the path $P$ is defined as the collection of all the iterated integrals of $P$:
\begin{equation}\label{sig}
  S(P)_{0,T}=(1,S(P)^1_{0,T},...,S(P)^d_{0,T},S(P)^{11}_{0,T},S(P)^{12}_{0,T},...),
\end{equation}
where the ``zeroth" term is 1 by convention, and the superscripts run along the set of all \emph{multi-indexes}
\begin{equation}\label{words}
  W = \{(i_1,...,i_k)|k\geq1,i_1,...,i_k\in\{1,...,d\}\}.
\end{equation}
And the $k^{th}$ level signature, a subset of $S(P)_{0,T}$, is defined to be the collection of all $k$-fold iterated integrals. Since the full $S(P)_{0,T}$ is an infinite series, in practice we usually truncate it at some level $m$. For example, in online handwriting recognition \cite{lai2017toward}\cite{xie2018learning} and writer identification \cite{yang2015chinese}\cite{yang2016deepwriterid}, $m$ is usually set between 2 and 5.
\subsubsection{Log Path Signature}
The LPS can be obtained by considering the tensor logarithm of the PS:
\begin{equation}\label{logsig}
  log S(P)_{0,T}=\sum_{n\geq1}\frac{(-1)^{n+1}}{n}(S(P)_{0,T}-1)^{\bigotimes n},
\end{equation}
where $\bigotimes$ is the tensor product. Chen's theorem \cite{chen1957integration} shows that the LPS can always be expressed using the Hall basis \cite{reutenauer2003free}; therefore, the LPS can be viewed as a more compact way to represent the PS or a dimensionality reduction.
\subsection{Offline Writer Identification Based on the LPS}
\subsubsection{Pathlet Extraction}
Although the offline handwritten text is not a time series, we can extract local \emph{pathlets} from the \emph{polygonized} handwriting contours as illustrated in Fig. 2 (a). First, the handwriting image is binarized using the Otsu's method, and the handwriting contours are extracted. Subsequently, a polygonization step, also called line approximation, is used to approximate each contour with a reduced number of points, with a tolerance approximation error $\epsilon$. The reason behind this is twofold. On the one hand, the raw contours are believed to have significant redundancy and noise, such as long straight lines and jagged edges owing to quantization, and can be well approximated with the structural information maintained. On the other hand, the computation can be much reduced. For example, by setting $\epsilon=1$, the number of points can be reduced by about 90\%.

After the above steps, we can simply trace a polygonized contour, starting from any point, and thus define the path. A pathlet is defined as a consecutive segment on the polygonized contour; using a sliding-window method, we can extract a large amount of pathlets and use the PS or LPS to describe their geometric properties. The pathlet size, i.e., the number of points in a pathlet, is an important parameter and should be appropriately chosen in order to cover expected local structures. Generally, the pathlet size should be inversely proportional to $\epsilon$.
\subsubsection{Feature Representation for Offline Handwriting}
An advantage of the LPS over the PS is that it has a lower dimensionality and is well distributed in the feature space, and hence, it is more suitable for the following clustering step to generate a feature codebook. Therefore we use the LPS in this paper. Nevertheless, the PS is also worthy of investigation, and we leave it for future work. Two feature normalization steps for LPS features are used. First, the length normalization method in \cite{lai2017online} is applied to deal with the length variation of the pathlets. Second, each feature dimension is rescaled to $[-1, 1]$. 
\begin{figure}[t]
\centering
\subfigure[]{\label{pathlet}\includegraphics[width=1.5in]{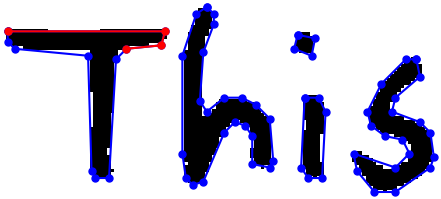}\hspace{0.3in}}
\subfigure[]{\label{pathletPair}\includegraphics[width=1.1in]{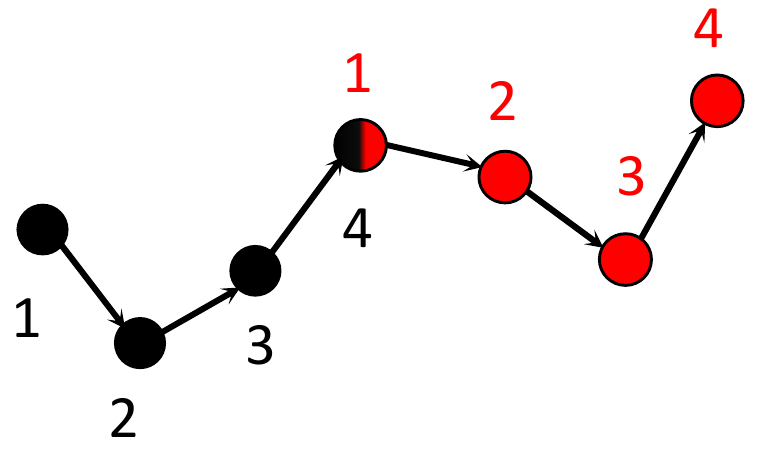}}\\
\subfigure[]{\label{2Dhist2}\includegraphics[width=2.2in]{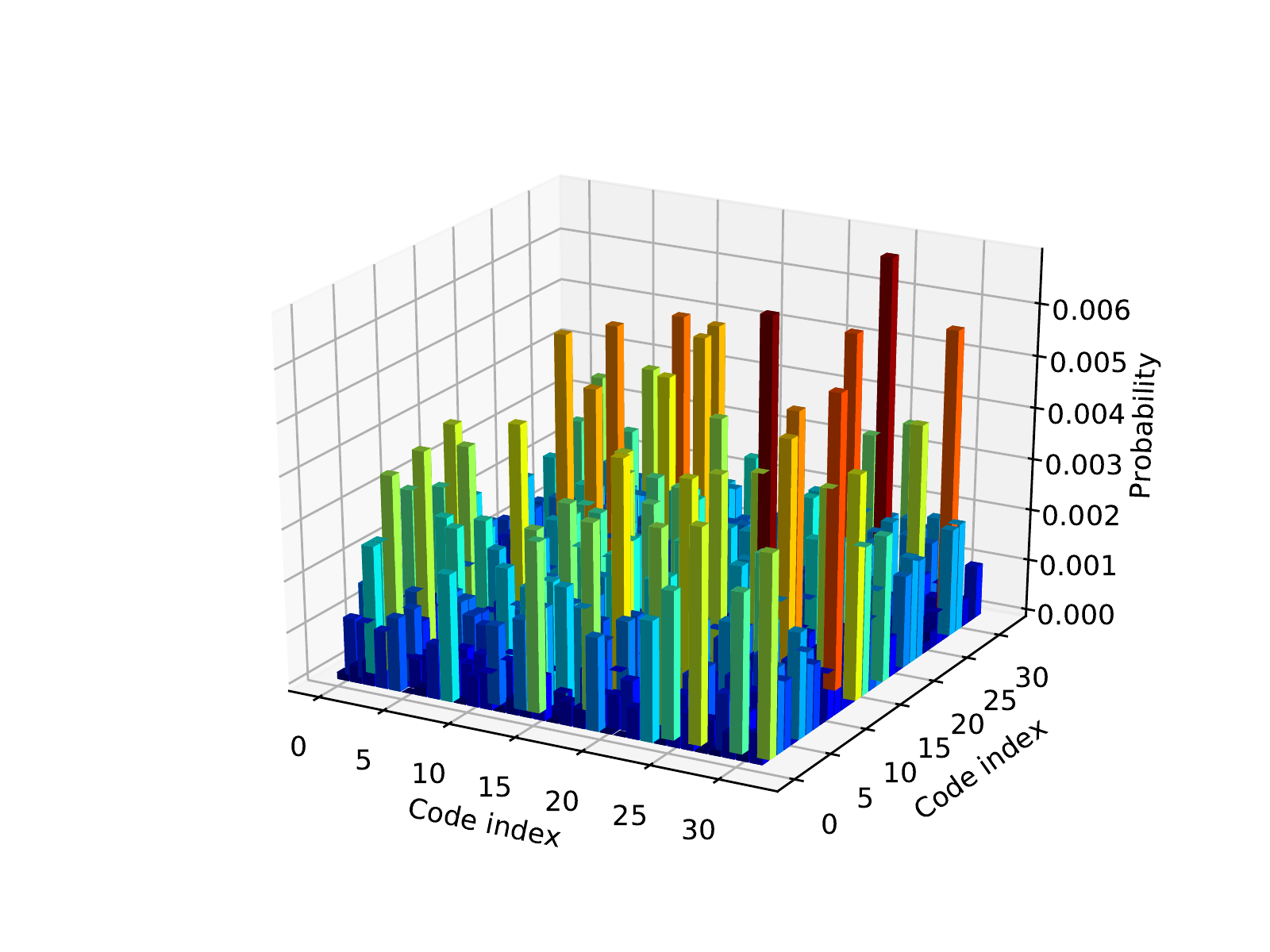}}
\caption{Construction of feature representation for offline handwriting. (a) Pathlets are extracted from the polygonized handwriting contours. (b) A pair of pathlets attached at a common end point. The joint LPS features of these pathlet pairs are extracted. (c) A feature matrix (i.e., 2D histogram) is computed from the joint LPS features based on a LPS feature codebook.}
\end{figure}

After LPS feature extraction, the bag-of-words method is used to construct the feature representation for offline handwriting owing to its simplicity. Specifically, the LPS features from training images are clustered using the k-means algorithm to obtain a feature codebook with $M$ elements. Given any new image, inspired by the Hinge method \cite{bulacu2007text}, we consider the joint LPS feature of a pair of pathlets attached at a common end point, as shown in Fig. 2 (b). Let $CB=\{\textbf{c}_1,\textbf{c}_2,...,\textbf{c}_M\}$ denote the codebook and $JL=\{[\textbf{f}_{11},\textbf{f}_{12}], [\textbf{f}_{21},\textbf{f}_{22}], ...,[\textbf{f}_{N1},\textbf{f}_{N2}]\}$ denote the extracted feature set, where $[\textbf{f}_{i1}, \textbf{f}_{i2}]$ is the joint LPS feature from the $i^{th}$ pathlet pair.
Based on $CB$ and $JL$, a $M\times M$ feature matrix $\textbf{FM}$ can be obtained as follows.
\begin{enumerate}
  \item Initialize $\textbf{FM}$ with zeros.
  \item For each $[\textbf{f}_{i1}, \textbf{f}_{i2}]\in JL$, find the nearest codes $\textbf{c}_{k_1}$ and $\textbf{c}_{k_2}$ to $\textbf{f}_{i1}$ and $\textbf{f}_{i2}$, respectively.
  \item $\textbf{FM}_{k_1k_2}=\textbf{FM}_{k_1k_2}+1$.
  \item Repeat steps 2 and 3 until all elements in $JL$ are visited.
  \item Normalize $\textbf{FM}$ to sum to 1.
\end{enumerate}
To measure the dissimilarity between two feature matrices $\textbf{U}$ and $\textbf{V}$, the Manhattan distance
\begin{equation}\label{L1distance}
  D_1(\textbf{U},\textbf{V}) = \sum_{i=1}^{M}\sum_{j=1}^{M}|\textbf{U}_{ij}-\textbf{V}_{ij}|,
\end{equation}
and the $\chi^2$ distance
\begin{equation}\label{X2distance}
  D_2(\textbf{U},\textbf{V}) = \sum_{i=1}^{M}\sum_{j=1}^{M}\frac{(\textbf{U}_{ij}-\textbf{V}_{ij})^2}{\textbf{U}_{ij}+\textbf{V}_{ij}},
\end{equation}
are used.
\subsubsection{Rotation Invariant Features}
\begin{figure}[t]
\centering
\subfigure[]{\label{rotInv1}\includegraphics[width=1.0in]{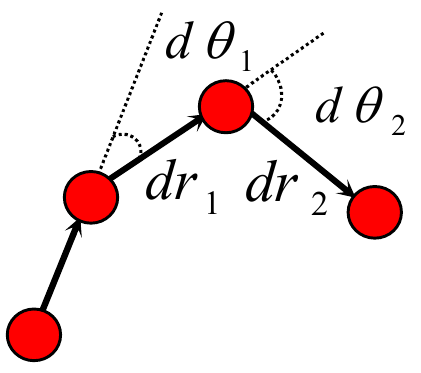}\hspace{0.3in}}
\subfigure[]{\label{rotInv2}\includegraphics[width=1.0in]{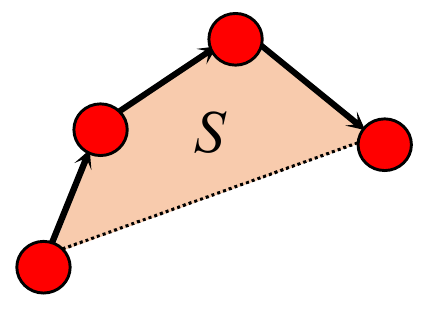}}
\caption{Based on the PS approach, rotation invariant features can be constructed. (a) A rotation invariant path, leading to rotation invariant PS and LPS features. (b) The $2^{nd}$ level term from the LPS, which is rotation invariant.}
\end{figure}
Based on the PS approach, rotation invariant features can also be constructed. A possible solution is to construct rotation invariant paths, e.g., paths in the polar coordinates. Another possible solution is to construct rotation invariant features directly \cite{diehl2013rotation} from the PS or LPS. For example, the $2^{nd}$ level term of the LPS is the Levy area enclosed by the path and the straight line connecting the end points. Fig. 3 illustrates the above two ideas. We leave these features for future work, as we do not focus on rotation invariant identification in the present study.
\section{Experiment Results and Analysis}
\subsection{Datasets and Performance Evaluation}
\begin{figure*}[t]
\centering
\subfigure[]{\label{effect_of_poly}\includegraphics[width=2.47in]{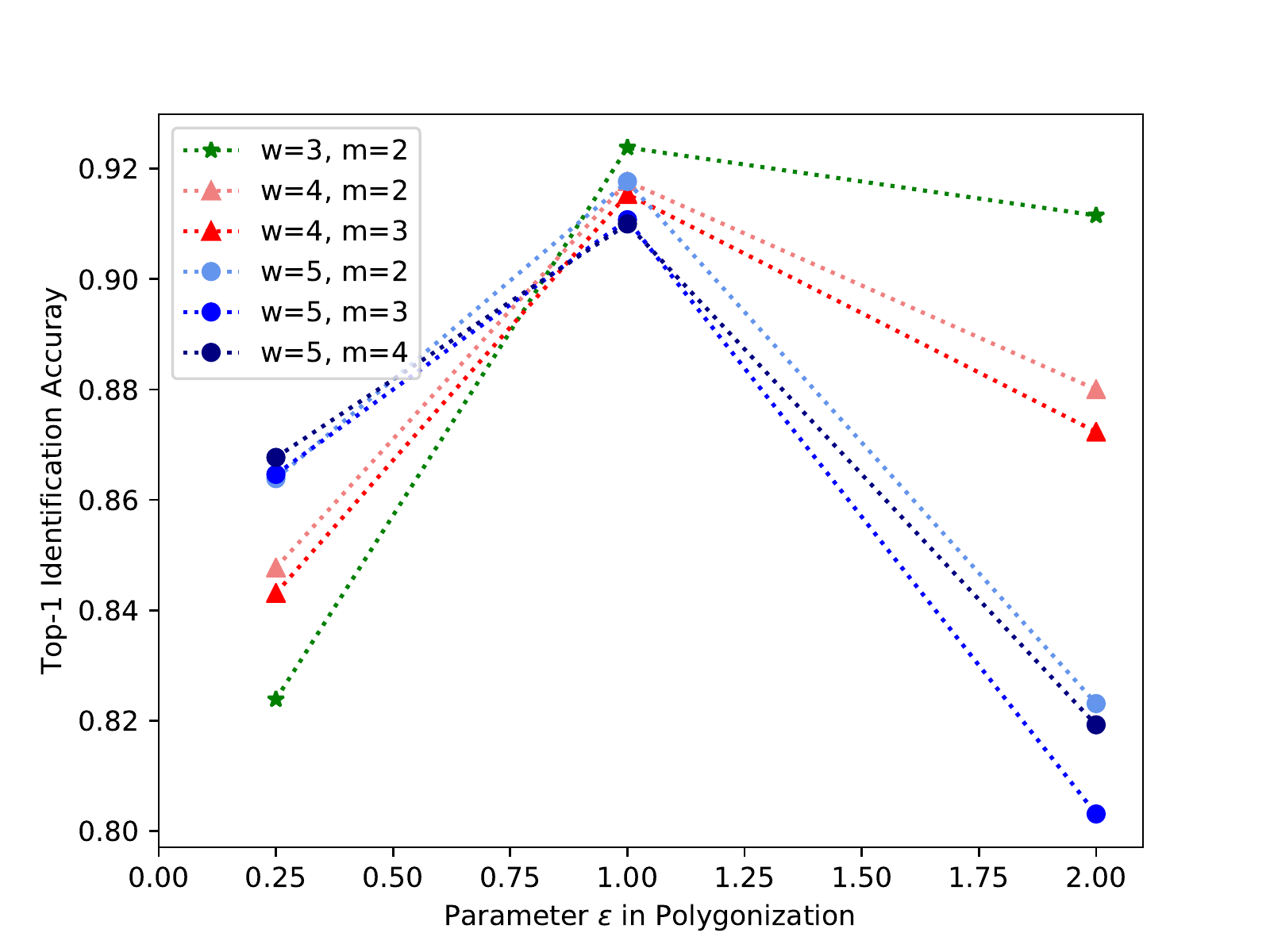}\hspace{0.3in}}
\subfigure[]{\label{effect_of_CB}\includegraphics[width=2.5in]{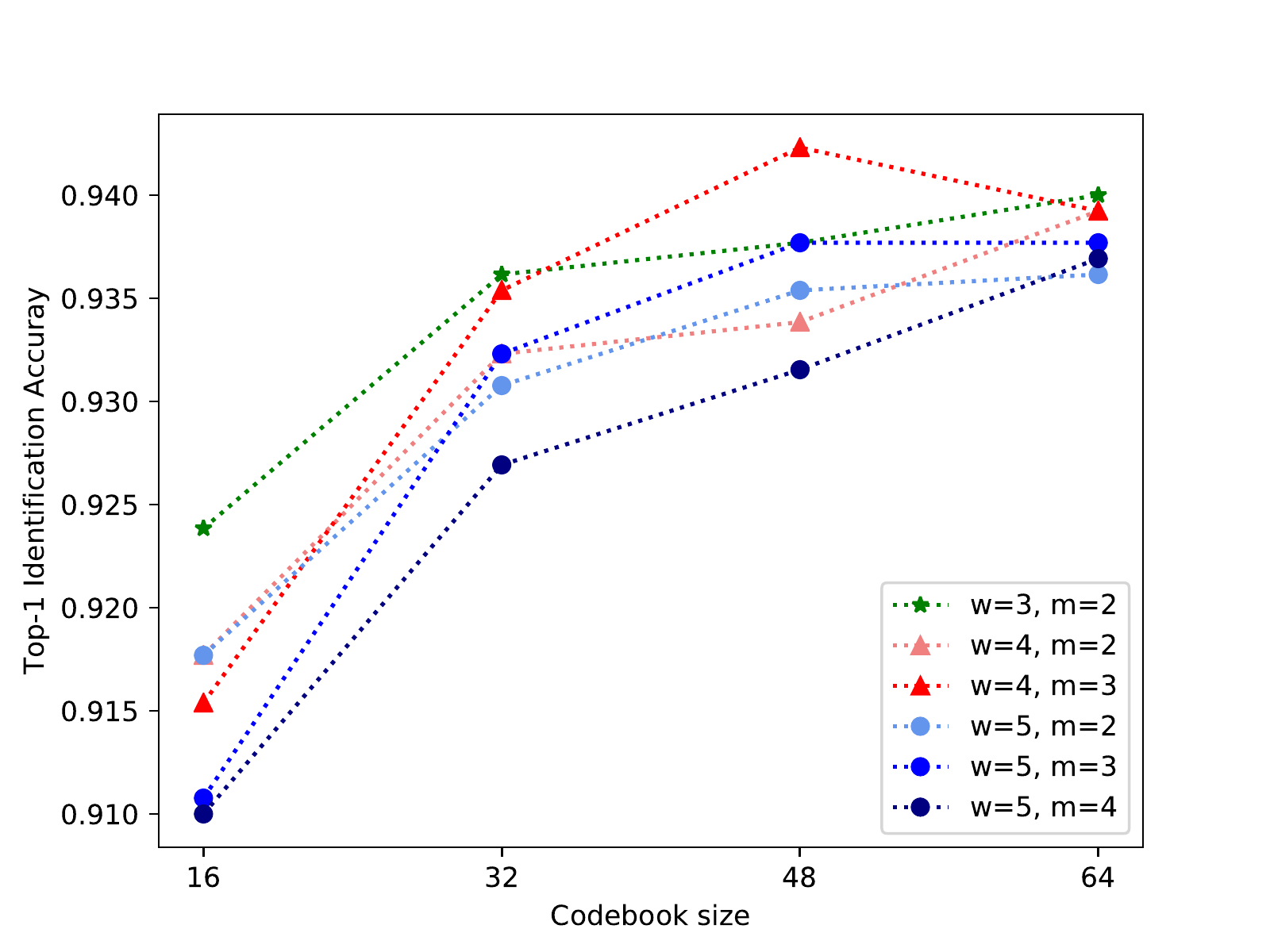}}\\
\caption{Effect of polygonization and codebook size. (a) Effect of $\epsilon$ on different pathlet parameter settings, with the codebook size fixed at 16. (b) Effect of codebook size on different pathlet parameter settings, with $\epsilon$ = 1.0.}
\end{figure*}
Our experiments are conducted on four benchmark datasets, namely IAM \cite{marti2002iam}, Firemaker \cite{bulacu2003writer}, CVL \cite{kleber2013cvl}, and ICDAR2013 writer identification contest dataset \cite{louloudis2013icdar}.

The IAM dataset is collected from 657 writers, and is modified in this work in a similar manner to that in \cite{bulacu2007text}. The modified IAM dataset used in our experiments has 650 writers, each having two handwritten documents.

The Firemaker dataset has 250 writers, each providing four pages of handwritten text. Following \cite{he2017beyond}, the page 1 and 4 are used in the experiments.

The CVL dataset consists of 1604 handwritten documents from 310 writers. There are 27 writers who provide seven samples (one German and six English) and 283 writers who provide five samples (one German and four English). Because there are empty samples in user 431, the first four English samples from 309 writers are used in the experiments.

The ICDAR2013 dataset contains 1000 handwritten documents from 250 writers. Each writer is asked to copy four pages of text in two languages (two in English and two in Greek). The entire ICDAR2013 dataset is used in our experiments, regardless of the language used.

To evaluate the identification performance, the ``leave-one-out" strategy is used: for each query document, we compute its distances to all other documents and sort the results in an ascending order. A correct hit is considered when at least one document of the same writer is included in the top N nearest neighbours. The ratio of the number of correct hits and the number of queries corresponds to the Top-N accuracy. In this paper we consider the Top-1 and Top-10 accuracies.
\subsection{Effect of Polygonization and Codebook Size}
The polygonization parameter $\epsilon$ controls the number of points that are removed from the raw contours, and has a direct effect on the performance. Therefore we first experiment with $\epsilon$ = 0.2, 1.0, and 2.0 on the IAM dataset. The pathlet size $w$ is chosen to be 3, 4, or 5, and the LPS truncation level $m$ is chosen to be $1<m<w$. The codebook size $M$ is fixed at 16. The $\chi^2$ distance is used for $\epsilon$ = 0.2; the Manhattan distance is used for $\epsilon$ = 1.0 and 2.0 and in all the following experiments.

Experiment results are shown in Fig. 4 (a). The best results are achieved at $\epsilon$ = 1.0. When $\epsilon$ = 0.2, a larger pathlet size can capture more meaningful local structures, therefore leads to an improved identification accuracy. When $\epsilon$ = 1.0, different pathlet sizes have similar performances. When $\epsilon$ = 2.0, a large pathlet size leads to a degraded performance. The reason may be twofold. First, a small codebook size is insufficient to represent the complex patterns of long pathlets. Second, the number of pathlets is much reduced when $\epsilon$ = 2.0, and therefore insufficient for stable representations of the documents. Therefore, the parameter $\epsilon$, codebook size $M$ and the amount of ink should be considered at the same time. In practice, we observe that $\epsilon=1.0$ is a good choice across different datasets.

To see the effect of codebook size, we fix $\epsilon=1.0$ and vary the codebook size $M$ from 16 to 32, 48, and 64.
The results are presented in Fig. 4 (b). We can see that, the performance is improved as the codebook size increases. Surprisingly, the pathlet setting $w=3, m=2$ steadily performs well. The setting $w=4$, $m=3$, $M=48$ achieves the best Top-1 accuracy of 94.23\%, which is among the best reported results of \emph{individual} state-of-the-art features (i.e., no feature combination is used).
\subsection{Results on Benchmark Datasets}
Two parameter settings, $w=3$, $m=2$, $M=32$ and $w=4$, $m=3$, $M=48$, are applied to several benchmark datasets, and the best results are reported in Table I. A better result is achieved with $w=3$, $m=2$, $M=32$ on the Firemaker dataset, whereas on the other three datasets, better results are achieved with $w=4$, $m=3$, $M=48$. The reason is that, in the Firemaker dataset, the page 4 is naturally written and contains much less ink than the page 1. Therefore the page 4 has insufficient pathlets to cover the $48\times 48$ feature matrix, as analyzed in the above experiments.
\begin{table}[tb]
\renewcommand{\arraystretch}{1.2}
\caption{Offline writer identification results on four benchmark datasets using the LPS feature.}
\label{table1}
\centering
\begin{tabular}{cccccc}
\hline
\multirow{2}{*}{Dataset}&\multirow{2}{*}{Top-1}&\multirow{2}{*}{Top-10}&\multicolumn{3}{c}{Parameter setting}\cr
\cline{4-6}
&&&$w$&$m$&$M$\cr
\hline
IAM&94.24&97.77&4&3&48\\
Firemaker&91&97&3&2&32\\
CVL&99.27&99.51&4&3&48\\
ICDAR2013&96.6&99.2&4&3&48\\
\hline
\end{tabular}
\end{table}


\begin{table}[tb]
\renewcommand{\arraystretch}{1.2}
\caption{Comparison with previous results on four benchmark datasets.}
\label{table2}
\centering
\begin{tabular}{ccccc}
\hline
Database&Method&Feature&Top-1&Top-10\cr
\hline
\multirow{6}{*}{IAM}&Bulacu et al.\cite{bulacu2007text}&Hybrid&89&97\\
&Wu et al.\cite{wu2014offline}&SDS+SOH&98.5&99.5\\
&He et al.\cite{he2015junction}&Junclets&83.3&94.4\\
&He et al.\cite{he2017beyond}&QuadHinge&93.2&96.5\\
&Khan et al.\cite{khan2017robust}$^*$&DCT features&97.2&-\\
&\textbf{Our method}&LPS&94.24&97.77\\
\hline
\multirow{6}{*}{Firemaker}&Bulacu et al.\cite{bulacu2007text}&Hybrid&83&95\\
&Wu et al.\cite{wu2014offline}&SDS+SOH&92.4&98.8\\
&He et al.\cite{he2015junction}&Junclets&80.6&94.0\\
&He et al.\cite{he2017beyond}&QuadHinge&92.2&97.2\\
&Khan et al.\cite{khan2017robust}$^*$&DCT features&89.47&-\\
&\textbf{Our method}&LPS&91&97\\
\hline
\multirow{4}{*}{CVL}&CS-UMD\cite{kleber2013cvl}&Graphems&97.9&99.4\\
&Wu et al.\cite{wu2014offline}&SDS+SOH&99.2&99.6\\
&Nicolaou et al.\cite{nicolaou2015sparse}&SRS-LBP&99.0&99.5\\
&\textbf{Our method}&LPS&99.27&99.51\\
\hline
\multirow{4}{*}{ICDAR2013}&CS-UMD-a\cite{louloudis2013icdar}&Graphems&95.1&99.1\\
&Wu et al.\cite{wu2014offline}&SDS+SOH&95.6&99.1\\
&Nicolaou et al.\cite{nicolaou2015sparse}&SRS-LBP&97.2&99.2\\
&\textbf{Our method}&LPS&96.6&99.2\\
\hline
\end{tabular}
\\\footnotesize{$^*$The ``leave-one-out" strategy is not used.}
\end{table}

\begin{figure*}[t]
\centerline{\includegraphics[width=6in]{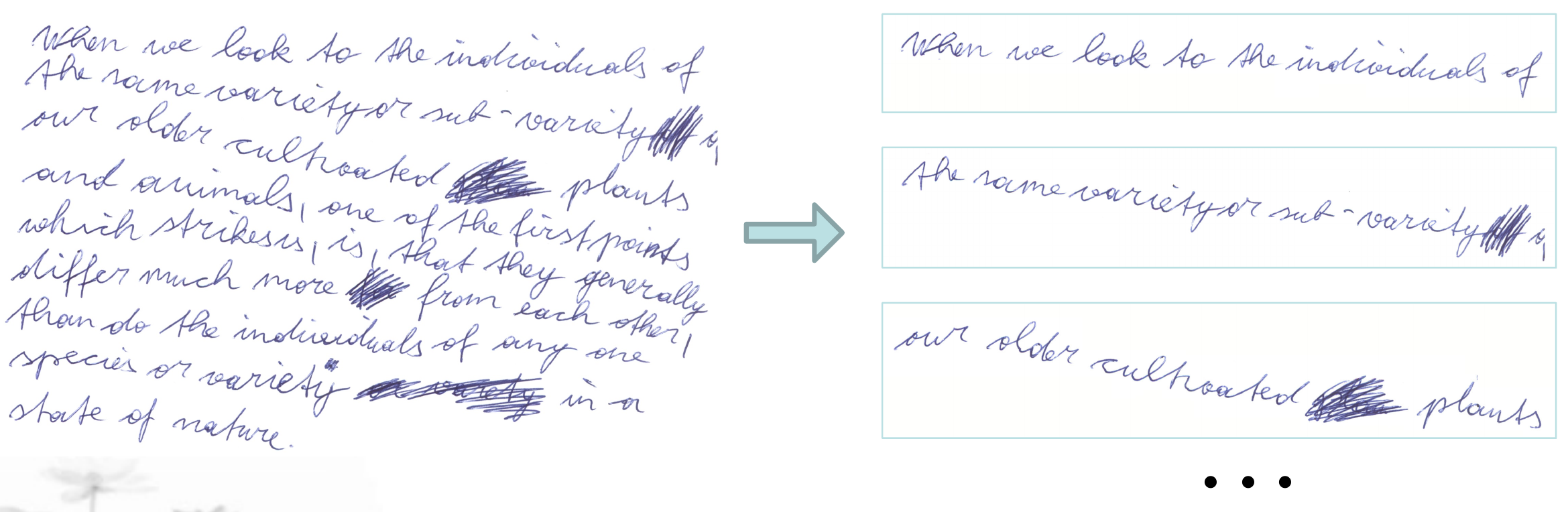}}
\caption{The fourth pages from the CVL dataset are cropped into text lines for line-level identification.}
\label{cvlCrop}
\end{figure*}

We compare the our results with previous results in Table II. Note that due to different experiment protocols in the literatures (such as whether the ``leave-one-out" strategy is used and how the dataset is used), some reported results should be treated differently. Our method achieves a very promising performance without any feature fusion or complex preprocessing such as segmentation. Thanks to the polygonization step and a small codebook size, our method is as fast as other contour-based methods, such as Hinge\cite{bulacu2007text}, at the same time. For example, the computation of the LPS feature is a magnitude faster than that of SIFT.
\subsection{Effect of Amount of Ink}
The amount of ink is important for the offline writer identification systems. To test how the performance of our method varies with amount of ink in \emph{query} documents, we conduct the following line-level identification experiment. We crop the fourth pages from the CVL dataset into text lines using the line projection method, with some necessary manual adjustments. Some examples are shown in Fig. 5. As there are at least four text lines within each document, we take the first four text lines from each document. When using a single text line as a query, the number of queries is $4\times 309=1236$; when using two text lines and three text lines, the numbers of queries are $C_4^2\times 309=1854$ and $C_4^3\times 309=1236$, respectively. The first three pages are used as templates, while the text lines are used as the test set. The ``leave-one-out" strategy is not applied here, and the setting $w=4$, $m=3$, $M=48$ is used. Experiment results are shown in Table III. Using only one single text line as a query, our method can achieve a top-1 accuracy of 95.31\%. Therefore our method is rather data efficient.

\begin{table}[tb]
\renewcommand{\arraystretch}{1.2}
\caption{Line-level writer identification on the CVL dataset.}
\label{table3}
\centering
\begin{tabular}{ccccccc}
\hline
Query sample&1 line&2 lines&3 lines&whole page\cr
\hline
Top-1 accuracy&95.31&98.44&99.03&99.35\\
\hline
\end{tabular}
\end{table}

\begin{table}[tb]
\renewcommand{\arraystretch}{1.2}
\caption{Performance of two parameter settings on the Firemaker dataset to see the effect of amount of ink.}
\label{table4}
\centering
\begin{tabular}{cccccc}
\hline
\multirow{2}{*}{Template sample} & \multirow{2}{*}{Top-1} & \multirow{2}{*}{Top-10} & \multicolumn{3}{c}{Parameter setting}\cr
\cline{4-6}
&&&$w$&$m$&$M$\\
\hline
Page 1&95.6&97.6&\multirow{2}{*}{3}&\multirow{2}{*}{2}&\multirow{2}{*}{32}\\
Page 4&86.4&96.4&&&\\
\hline
Page 1&96.8&98.8&\multirow{2}{*}{4}&\multirow{2}{*}{3}&\multirow{2}{*}{48}\\
Page 4&82.8&96.8&&&\\
\hline
\end{tabular}
\end{table}

We should point out that the templates should be representative of the writers. For example, three pages of text are used as templates to achieve the results in Table III. If the amount of ink in \emph{templates} is insufficient, statistical methods, including ours, would have a degraded performance. Indeed, this is why the Firemaker dataset prefers the setting $w=3$, $m=2$, $M=32$ rather than $w=4$, $m=3$, $M=48$. In Table IV we give the results of the two settings on the Firemaker dataset, and consider whether pages 1 or pages 4 are used as templates. The setting $w=3$, $m=2$, $M=32$ is better when using pages 4 as templates, whereas the setting $w=4$, $m=3$, $M=48$ is better when using pages 1 as templates. This is because the amount of ink in pages 4 is much less than that in pages 1. It is an important research direction to improve the performance when the amount of ink in templates is insufficient.
\section{Conclusion}
In this paper, we propose a novel set of features for offline writer identification based on the PS approach. The PS provides a principled way to express information contained in a path, such as orientation and curvature, and has achieved significant success in time series description. By extracting local pathlets from offline handwriting images, the PS can also to characterize the offline handwriting style. A codebook method based on the LPS---a more compact way to express the PS---is used in this work and shows competitive results on several benchmark datasets. Investigations of the PS, rotation invariant features and feature fusion are leaved for future work. Furthermore, how to improve the performance when no sufficient amount of ink is available is an important research direction.




%


\end{document}